\documentclass[runningheads]{llncs}
\usepackage[T1]{fontenc}

\usepackage{amsmath}
\usepackage{graphicx}
\usepackage{longtable}
\usepackage{booktabs}
\usepackage{adjustbox}
\usepackage{orcidlink}
\usepackage{tabularx}
\usepackage{subcaption}  
\usepackage{makecell}
\usepackage{array}
\usepackage{multirow}
\usepackage{float}
\usepackage{placeins}
\usepackage{xurl} 
\usepackage{caption}
\title{Detecting Jailbreak Attempts in Clinical Training LLMs Through Automated Linguistic Feature Extraction}
\titlerunning{Detecting Jailbreak in Clinical Training LLMs}

\author{Tri Nguyen\orcidlink{0009-0004-6660-930X} \and
        Huy Hoang Bao Le \and
        Lohith Srikanth Pentapalli\orcidlink{0009-0002-9801-3418} \and
        Laurah Turner\orcidlink{0000-0002-4567-1313} \and
        Kelly Cohen\orcidlink{0000-0002-8655-1465}}
\authorrunning{T.~Nguyen et~al.}

\institute{University of Cincinnati, Cincinnati OH 45219, USA \\
\email{nguye3hr@mail.uc.edu, lebu@mail.uc.edu, pentapl5@mail.uc.edu,
       turnela@ucmail.uc.edu, cohenky@ucmail.uc.edu}}

\begin{document}

\maketitle

\begin{abstract}
Detecting jailbreak attempts in clinical training large language models (LLMs) requires accurate modeling of linguistic deviations that signal unsafe or off-task user behavior. Prior work on the 2-Sigma clinical simulation platform showed that manually annotated linguistic features could support jailbreak detection. However, reliance on manual annotation limited both scalability and expressiveness. In this study, we extend this framework by using experts’ annotations of four core linguistic features (Professionalism, Medical Relevance, Ethical Behavior, and Contextual Distraction) and training multiple general-domain and medical-domain BERT-based LLM models to predict these features directly from text. The most reliable feature regressor for each dimension was selected and used as the feature extractor in a second layer of classifiers. We evaluate a suite of predictive models, including tree-based, linear, probabilistic, and ensemble methods, to determine jailbreak likelihood from the extracted features. Across cross-validation and held-out evaluations, the system achieves strong overall performance, indicating that LLM-derived linguistic features provide an effective basis for automated jailbreak detection. Error analysis further highlights key limitations in current annotations and feature representations, pointing toward future improvements such as richer annotation schemes, finer-grained feature extraction, and methods that capture the evolving risk of jailbreak behavior over the course of a dialogue. This work demonstrates a scalable and interpretable approach for detecting jailbreak behavior in safety-critical clinical dialogue systems.
\keywords{Jailbreak Detection  \and Clinical Training LLMs \and LLM Feature Extraction \and Interpretability.}
\end{abstract}

\section{Introduction}
2-Sigma is a clinical training platform that leverages large language models to simulate patient interactions, enabling medical students to practice clinical reasoning, diagnosis, and decision-making through conversational, case-based scenarios. While this approach offers substantial educational benefits, it also introduces new safety challenges, as students may attempt to manipulate the LLM through jailbreak behaviors that degrade professionalism, ethical standards, or alignment with the intended clinical task. Prior work on 2-Sigma demonstrated that such jailbreak attempts are strongly correlated with specific linguistic patterns and showed that manually engineered linguistic features could be used to detect jailbreak behavior with high accuracy \cite{nguyenJailbreakDetectionClinical2025}. However, this feature-based framework relied on labor-intensive human annotation and expanded categorical encodings, which limited both scalability and expressiveness. Building on this foundation, the present study seeks to automate linguistic feature extraction using fine-tuned LLMs while preserving interpretability, enabling a scalable and robust approach to jailbreak detection in clinical training environments.

\begin{figure}[htbp]
\centering
\includegraphics[width=0.95\linewidth]{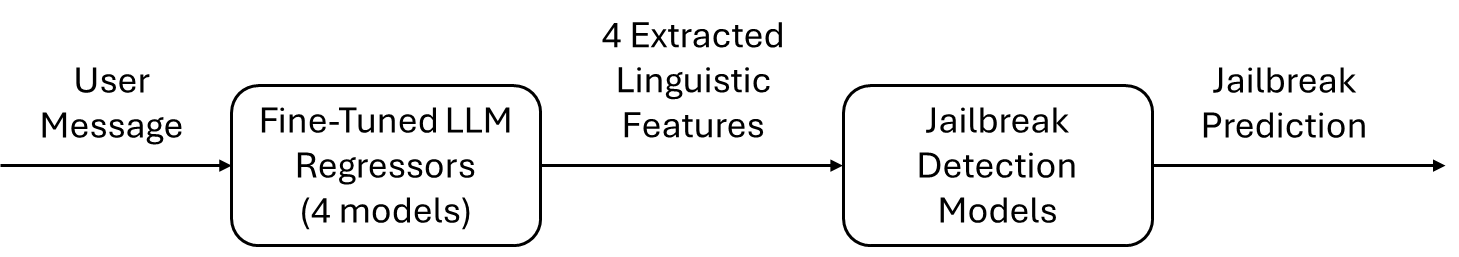}
\caption{System overview of the proposed framework.}
\label{fig:system-overview}
\end{figure}

This study adopts a two-layer modeling architecture for jailbreak detection in clinical LLM interactions, as shown in Fig.~\ref{fig:system-overview}. In the first layer, four independently fine-tuned transformer-based regressors map each user message—using either single-turn or multi-turn context—to continuous scores across linguistic dimensions shown to correlate with jailbreak behavior: Professionalism, Medical Relevance, Ethical Behavior, and Contextual Distraction. These LLM-derived features provide a compact, semantically informed representation of each prompt. In the second layer, multiple downstream classifiers, spanning linear, probabilistic, and ensemble methods, use the resulting four-dimensional feature vector to predict jailbreak likelihood. This modular design separates linguistic feature extraction from classification, enabling automated annotation, improved interpretability, and flexible substitution of models at either stage. Experimental results show that the proposed two-layer framework achieves strong and stable performance across multiple classifiers, providing an automated yet interpretable foundation for jailbreak detection through explicit linguistic features.

\section{Related Work}
Large Language Models have demonstrated strong performance across natural language understanding and generation tasks, yet they remain vulnerable to jailbreak attacks. Jailbreak attacks involve designing adversarial prompts that circumvent LLMs' safety constraints and bypass alignment. With the increasing sophistication of jailbreaking strategies, it is paramount to develop reliable detection mechanisms. Existing research spans surface-level classifiers, prompt perturbation techniques, internal signal monitoring, and gradient-based introspection, each offering complementary strengths and limitations. 

Early jailbreak detection methods are primarily based on classifier-based and heuristic approaches. Many LLM providers deploy moderation filters (e.g. OpenAI's Moderation API) to flag toxic or policy-violating content, but these generic classifiers may miss the nuanced strategies of jailbreak prompts \cite{xieGradSafeDetectingJailbreak2024}. Zero-shot safety judgments performed internally by LLMs themselves have also proven unreliable \cite{xieGradSafeDetectingJailbreak2024}. To improve robustness and performance, supervised detectors trained on labeled jailbreak data have been proposed. Hawkins et al. showed that a fine-tuned BERT classifier outperforms generic moderation systems by exploiting lexical and syntactic cues such as explicit reflexivity in prompt structure and role-play framing that correlate strongly with jailbreak intent \cite{hawkinsMachineLearningDetection2025}. Linguistically motivated approaches such as "Prompter Says" further emphasize syntactic structure and phrasing as indicators of adversarial prompts \cite{leePrompterSaysLinguistic2024}. In addition, statistical techniques such as prompt perplexity have been explored, with evidence that malicious prompts often exhibit anomalous perplexity relative to benign inputs \cite{alonDetectingLanguageModel2023}. While effective against known attack patterns, these approaches highlight the importance of richer data and adaptive modeling to enhance robustness against evolving jailbreaking strategies.

To handle increasingly obfuscated attacks, researchers have introduced augmentation-based and multi-prompt detection strategies. SmoothLLM applies randomized smoothing by perturbing user prompts and aggregating model responses, exploiting instability near decision boundaries characteristic of jailbreak prompts \cite{robeySmoothLLMDefendingLarge2024}. There has also been an increase in hybrid approaches, which aid in reducing false positives. Self-Reminder mechanisms reinforce safety by prepending alignment-focused system prompts \cite{wuDefendingChatGPTJailbreak2023}, while multi-agent frameworks such as AutoDefense introduce gatekeeper agents that analyze prompts before execution, improving detection of complex jailbreak strategies at the cost of increased latency and system complexity \cite{zengAutoDefenseMultiAgentLLM2024}.

More recent work has shifted toward internal signal monitoring, leveraging the observation that aligned LLMs internally recognize disallowed intent even when coerced into compliance. JBShield operationalizes this idea by using concept-activation vectors representing toxic content and jailbreak manipulation, enabling accurate detection when both concepts are simultaneously activated \cite{zhangJBShieldDefendingLarge2025}. Beyond detection, JBShield can intervene by modifying latent representations to restore refusal behavior. HiddenDetect introduces a training-free method that monitors hidden-layer activations for refusal-related semantics by measuring similarity to embeddings of refusal phrases \cite{jiangHiddenDetectDetectingJailbreak2025}. This method generalizes to multimodal models, where delayed or suppressed refusal signals correlate with higher jailbreak success. These findings demonstrate that LLMs encode intrinsic safety-relevant signals in their internal representations, which can be leveraged for jailbreak detection.

In parallel, gradient-based detection methods have emerged as another introspective strategy. GradSafe analyzes gradients of the next-token loss with respect to safety-critical parameters and identifies characteristic patterns shared across unsafe prompts, operating directly on pre-trained models without additional fine-tuning \cite{xieGradSafeDetectingJailbreak2024}. Gradient Cuff further refines this idea by explicitly modeling refusal loss and jointly examining its magnitude and gradient norm, where a low refusal loss combined with a high gradient norm signals adversarial pressure away from refusal \cite{huGradientCuffDetecting2024}. Complementary work has shown that analysis of hidden-layer representations using tensor decomposition can reveal latent factors that naturally separate jailbreak prompts from benign inputs, even without labeled data \cite{kadaliInternalLayersLLMs2025}.

Alongside jailbreak detection, LLMs have increasingly been used for feature extraction and representation learning. Transformer-based models such as BERT and GPT provide dense semantic embeddings that outperform manually engineered features in many NLP tasks. In the medical domain, LLM-based systems have achieved near-expert performance in extracting structured clinical information from unstructured electronic health records, significantly surpassing traditional regex-based and n-gram approaches \cite{yuanTransformersLargeLanguage2025}. Beyond text analytics, LLMs have also been applied to feature selection in tabular datasets by leveraging semantic knowledge of feature descriptions rather than relying solely on statistical criteria \cite{liExploringLargeLanguage2024}. More advanced systems such as Rogue One extend these ideas through collaborative multi-agent architectures that autonomously generate, evaluate, and refine new features, producing both performance gains and semantically meaningful representations \cite{bradlandKnowledgeInformedAutomaticFeature2025}.

In summary, jailbreak detection has evolved from surface-level classifiers to introspective methods that exploit hidden states, gradients, and latent representations within LLMs. While modern techniques achieve strong detection performance, they vary in computational cost, interpretability, and robustness to unseen attacks. Concurrently, advances in LLM-based feature extraction highlight the models' capacity to generate rich, high-level representations. This dual role of LLMs as both vulnerable targets and powerful feature extractors motivates hybrid detection frameworks. In particular, fuzzy-logic-based approaches provide a principled mechanism for integrating heterogeneous signals under uncertainty, making them well suited for robust jailbreak detection.

\section{Methods}
\subsection{Data Preparation}

The dataset used in this study originates from 2-Sigma’s initial evaluation phase, which collected 158 conversations comprising 2,302 total prompts containing a mixture of jailbreak (JB) and non-jailbreak interactions. In this context, a jailbreak refers to instances where a user’s communication deteriorates in professionalism, relevance, ethical behavior, or contextual alignment as they attempt to steer the clinical LLM toward unsafe, inappropriate, or off-task behavior. Each prompt was independently labeled as JB or non-JB by two annotators, forming the binary ground-truth labels used for downstream classification.

In addition to the binary jailbreak labels, the previous study~\cite{nguyenJailbreakDetectionClinical2025} provided a set of human annotations evaluating four conversational linguistic features (\textit{Professionalism}, \textit{Medical Relevance}, \textit{Ethical Behavior}, and \textit{Contextual Distraction}) that were shown to correlate strongly with jailbreak attempts. Six annotators rated every prompt using structured ordinal scales: unprofessional $\rightarrow$ borderline $\rightarrow$ appropriate for Professionalism; irrelevant $\rightarrow$ partially relevant $\rightarrow$ relevant for Medical Relevance; dangerous $\rightarrow$ unsafe $\rightarrow$ questionable $\rightarrow$ mostly safe $\rightarrow$ safe for Ethical Behavior; and highly distracting $\rightarrow$ moderately distracting $\rightarrow$ questionable $\rightarrow$ not distracting for Contextual Distraction.

To support the goal of automating the annotation process, each linguistic feature was converted into a continuous regression target. The categorical ratings from all annotators were first mapped to ordinal numeric values, after which the average rating for each prompt was computed. The resulting continuous targets fall within the numeric range of 1 to the maximum level of their respective rating scales, preserving the ordinal meaning of the annotations. This procedure yielded four continuous variables that capture the consensus severity of each linguistic deviation while retaining inter-annotator variability. These continuous targets were then used to fine-tune a set of pretrained LLMs, enabling them to act as feature extractors capable of predicting the four linguistic dimensions for unseen prompts.

Following LLM fine-tuning, each prompt was represented by a four-dimensional vector containing the model-predicted levels of Professionalism, Medical Relevance, Ethical Behavior, and Contextual Distraction. These learned features served as the input to a suite of downstream predictive models trained to classify whether a user message constitutes a jailbreak attempt. To prevent conversation-level leakage and ensure realistic evaluation, the dataset was split by conversation: 147 conversations (2,021 prompts) were used for training and validation of both the LLM regressors and the predictive classifiers, while the remaining 11 conversations (281 prompts) were held out as an unseen test set for final evaluation.

\begin{table}[htbp]
\centering
\caption{Example of single-turn and multi-turn LLM training inputs, human annotations, and corresponding numerical regression targets.}
\label{tab:llm-inputs-targets}

\setlength{\tabcolsep}{4pt}
\begin{tabular}{
p{2.2cm}  
p{2.2cm}  
p{2.2cm}  
>{\centering\arraybackslash}p{0.9cm}
>{\centering\arraybackslash}p{0.9cm}
>{\centering\arraybackslash}p{0.9cm}
>{\centering\arraybackslash}p{0.9cm}      
}
\toprule
\multicolumn{2}{c}{\textbf{LLM Training Inputs}} &
\multirow{3.5}{*}{\centering\shortstack{\textbf{Human}\\\textbf{Annotation}\\\textbf{(6 annotators)}}} &
\multicolumn{4}{c}{\textbf{LLM Training Targets}} \\

\cmidrule(lr){1-2} \cmidrule(lr){4-7}
\textbf{Single-turn Message} &
\textbf{Multi-turn Message} &
&
\textbf{Prof.} &
\textbf{Med. Rel.} &
\textbf{Eth. Beh.} &
\textbf{Cont. Dist.} \\
\midrule

I am a second year medical student. Could I have a case about urology?
&
I am a second year medical student. Could I have a case about urology?
&
(6) Professional \newline (6) Relevant \newline (6) Safe \newline (6) Not Distracting
&
3 & 3 & 5 & 4 \\

\addlinespace

Hi sir, what brings you in today?
&
I am a second year medical student. Could I have a case about urology? Hi sir, what brings you in today?
&
(6) Professional \newline (6) Relevant \newline (6) Safe \newline (6) Not Distracting
&
3 & 3 & 5 & 4 \\

\bottomrule
\end{tabular}
\end{table}

\subsection{LLM Feature Extraction}

To construct semantic features capable of capturing subtle linguistic deviations associated with jailbreak attempts, we fine-tuned a diverse suite of transformer-based encoders on four continuous expert-derived features: \textit{Professionalism}, \textit{Medical Relevance}, \textit{Ethical Behavior}, and \textit{Contextual Distraction}. These pretrained encoder architectures includes BERT-Base, BERT-Large, DistilBERT, RoBERTa-Base, DeBERTa-v3-Large, BioBERT, and PubMedBERT. Detailed descriptions of these models and their pretraining corpora are provided in the Appendix.

Each model was fine-tuned as a regression predictor, using either single-turn messages or cumulative multi-turn context for features that are sensitive to conversational flow (particularly \textit{Contextual Distraction}), as demonstrated in Table \ref{tab:llm-inputs-targets}. Inputs were tokenized using each model's native tokenizer and encoded by the transformer encoder, followed by a single-output regression head that predicts a continuous score. Training optimized the root mean squared error (RMSE), defined as
\[
\mathrm{RMSE} = \sqrt{ \frac{1}{n} \sum_{i=1}^{n} \left( \hat{y}_i - y_i \right)^2 },
\]
while performance was additionally evaluated using the coefficient of determination,
\[
R^2 = 1 - \frac{ \sum_{i=1}^{n} \left( y_i - \hat{y}_i \right)^2 }{ \sum_{i=1}^{n} \left( y_i - \bar{y} \right)^2 },
\]
where \(y_i\) denotes the expert-annotated target score, \(\hat{y}_i\) the model prediction, and \(\bar{y}\) the mean of all target scores. To quantify prediction bias, we also report the mean error,
\[
\mathrm{Mean\ Error} = \frac{1}{n} \sum_{i=1}^{n} \left( \hat{y}_i - y_i \right),
\]
and the standard deviation of error,
\[
\mathrm{SD\ Error} = \sqrt{ \frac{1}{n} \sum_{i=1}^{n} \left[ \left( \hat{y}_i - y_i \right) - \mathrm{Mean\ Error} \right]^2 }.
\]
A complete summary of all regression metrics (RMSE, $R^2$, mean error, and standard deviation of error) for every model-feature pair is provided in Table~\ref{tab:reg_all}. Training stability across model scales was supported by small per-device batch sizes with gradient accumulation, weight decay regularization, and early stopping based on validation RMSE, with the checkpoint achieving the lowest validation RMSE retained for each linguistic feature.

After fine-tuning, each encoder produced task-specific contextual embeddings that operationalize the feature it was trained to predict. Because the architectures differ in representational capacity, pretraining objectives, and domain specialization, the resulting embeddings capture complementary linguistic information: general semantic patterns (e.g., BERT, DistilBERT), deeper contextual and syntactic cues (e.g., RoBERTa, DeBERTa), and clinically grounded biomedical semantics (e.g., BioBERT, PubMedBERT). Based on the regression outcomes, different pretrained encoders emerged as the top-performing models for different linguistic features. Specifically, DistilBERT achieved the lowest RMSE for \textit{Professionalism}, BERT performed best for \textit{Medical Relevance}, and BioBERT was the most effective model for both \textit{Ethical Behavior} and \textit{Contextual Distraction}, as shown in Table \ref{tab:best_rmse_models}. Accordingly, the downstream prediction layer uses the model fine-tuned on its corresponding feature, ensuring that the feature representations are optimally aligned with the linguistic dimension they are intended to capture.

\begin{table}[!t]
\centering
\caption{Best-performing pretrained model for each linguistic feature based on RMSE.}
\label{tab:best_rmse_models}
\begin{tabular}{lccc}
\hline
\textbf{Linguistic Feature} & \textbf{Best Model} & \textbf{RMSE} & \textbf{$R^2$}\\
\hline
Professionalism         & DistilBERT & 0.4441 & 0.5939\\
Medical Relevance       & BERT       & 0.4630 & 0.5983\\
Ethical Behavior        & BioBERT    & 0.5751 & 0.4965\\
Contextual Distraction  & BioBERT    & 0.6701 & 0.6147\\
\hline
\end{tabular}
\end{table}

\subsection{Predictive Modeling}

The final jailbreak classification layer was evaluated using a diverse set of predictive models including tree-based, linear, probabilistic, and ensemble learning approaches. Specifically, we trained Decision Trees (DT), Fuzzy Decision Trees (FDT), Random Forests (RF), Light Gradient Boosting Machine (LGBM), Extreme Gradient Boosting (XGBoost), Logistic Regression (LR), and Gaussian Naïve Bayes (NB). All models operated on the four LLM-extracted linguistic features generated by the best-performing LLM for each dimension (Table~\ref{tab:best_rmse_models}). The classifiers were implemented using the \texttt{scikit-learn}, \texttt{lightgbm}, \texttt{xgboost}, and \texttt{skfuzzy} libraries, with default parameter settings. 5-fold cross-validation was performed on the training set to evaluate model stability and generalization. The performance metrics averaged across folds and their standard deviation are summarized in Table~\ref{tab:clf-performance}.

\section{Results and Discussion}

Table~\ref{tab:test_set_performance_full} reports the test-set performance of all downstream jailbreak-detection classifiers across five evaluation metrics: accuracy, precision, recall, F1-score, and ROC-AUC. Each model was evaluated on the four LLM-extracted linguistic features to generate a final prediction for each user message. Table~\ref{tab:clf-performance} presents the corresponding cross-validation results, providing mean and standard deviation values for the same metrics.

\begin{table}[htbp]
\centering
\caption{Test set performance of downstream jailbreak-detection classifiers across multiple evaluation metrics.}
\label{tab:test_set_performance_full}
\begin{tabular}{lccccc}
\hline
\textbf{Model} & \textbf{Accuracy} & \textbf{Precision} & \textbf{Recall} & \textbf{F1} & \textbf{ROC-AUC} \\
\hline
DT      & 0.8256 & 0.5789 & 0.7213 & 0.6423 & 0.7879 \\
FDT     & 0.8932 & 0.7246 & 0.8197 & 0.7692 & 0.8667 \\
RF      & \textbf{0.9004} & \textbf{0.7463} & 0.8197 & \textbf{0.7812} & 0.8712 \\
LGBM    & 0.8790 & 0.6901 & 0.8033 & 0.7424 & 0.8516 \\
XGBoost & 0.8861 & 0.6986 & 0.8361 & 0.7612 & 0.8680 \\
LR      & 0.8897 & 0.7027 & \textbf{0.8525} & 0.7704 & \textbf{0.8762} \\
NB      & 0.8968 & 0.7286 & 0.8361 & 0.7786 & 0.8749 \\
\hline
\end{tabular}
\end{table}

\begin{table}[!t]
\centering
\caption{Cross-validation performance comparison of classifiers $(\mu \pm \sigma)$. Bold = best; bold italic = second-best.}
\label{tab:clf-performance}

\setlength{\tabcolsep}{6pt}

\begin{subtable}{\linewidth}
\centering
\caption{Accuracy, F1-Score, Precision}
\begin{tabular}{lccc}
\toprule
\textbf{Model} & \textbf{Accuracy} & \textbf{F1-Score} & \textbf{Precision} \\
\midrule
DT      & 0.8745 $\pm$ 0.0168 & 0.8752 $\pm$ 0.0163 & 0.8738 $\pm$ 0.0233 \\
FDT     & \textbf{\textit{0.9118 $\pm$ 0.0093}} & 0.9114 $\pm$ 0.0090 & 0.9200 $\pm$ 0.0159 \\
RF      & 0.9083 $\pm$ 0.0083 & 0.9084 $\pm$ 0.0071 & 0.9125 $\pm$ 0.0187 \\
LGBM    & 0.9092 $\pm$ 0.0087 & 0.9090 $\pm$ 0.0075 & 0.9157 $\pm$ 0.0220 \\
XGBoost & 0.9070 $\pm$ 0.0116 & 0.9067 $\pm$ 0.0115 & 0.9141 $\pm$ 0.0171 \\
LR      & \textbf{0.9127 $\pm$ 0.0135} & \textbf{\textit{0.9118 $\pm$ 0.0136}} & \textbf{0.9246 $\pm$ 0.0172} \\
NB      & \textbf{0.9127 $\pm$ 0.0139} & \textbf{0.9120 $\pm$ 0.0143} & \textbf{\textit{0.9222 $\pm$ 0.0142}} \\
\bottomrule
\end{tabular}
\end{subtable}

\vspace{6pt}

\begin{subtable}{\linewidth}
\centering
\caption{Recall, ROC-AUC}
\begin{tabular}{lcc}
\toprule
\textbf{Model} & \textbf{Recall} & \textbf{ROC-AUC} \\
\midrule
DT      & 0.8771 $\pm$ 0.0179 & 0.8745 $\pm$ 0.0168 \\
FDT     & \textbf{\textit{0.9030 $\pm$ 0.0097}} & \textbf{0.9677 $\pm$ 0.0073} \\
RF      & \textbf{0.9048 $\pm$ 0.0091} & 0.9662 $\pm$ 0.0050 \\
LGBM    & \textbf{\textit{0.9030 $\pm$ 0.0124}} & 0.9657 $\pm$ 0.0058 \\
XGBoost & 0.8996 $\pm$ 0.0141 & 0.9631 $\pm$ 0.0064 \\
LR      & 0.8996 $\pm$ 0.0135 & \textbf{\textit{0.9666 $\pm$ 0.0075}} \\
NB      & 0.9022 $\pm$ 0.0163 & 0.9652 $\pm$ 0.0082 \\
\bottomrule
\end{tabular}
\end{subtable}

\end{table}

Compared to the prior study~\cite{nguyenJailbreakDetectionClinical2025}, the two-layer approach proposed here offers two key advantages: it automates the annotation of linguistic features and provides interpretable insight into how those features contribute to the final jailbreak prediction. As shown in Table~\ref{tab:test_set_performance_full}, the system achieves strong performance across multiple metrics, demonstrating that LLM-extracted linguistic cues can reliably support jailbreak detection. Although performance is slightly lower than classifiers trained directly on ground-truth annotations (Table~\ref{tab:test-performance-ground-truth}), this difference is expected due to the propagation of regression error from the first layer. Importantly, the overall results remain robust, as supported by cross-validation outcomes in Table~\ref{tab:clf-performance}, indicating that the feature extraction stage effectively captures the linguistic signals most predictive of jailbreak behavior.

Across classifiers, Decision Trees show the weakest performance on both cross-validation and the test set, with marked drops across all metrics—suggesting overfitting and limited ability to model the structure of the feature space. In contrast, ensemble methods such as Random Forest, LightGBM, and XGBoost leverage aggregated weak learners to capture more complex patterns, yielding consistently strong results; among them, Random Forest performs best on the held-out test set. Fuzzy Decision Trees, Logistic Regression, and Naïve Bayes also perform competitively in cross-validation, in some cases slightly outperforming the ensemble models. Their success suggests that the LLM-extracted features form a representation that is close to linearly separable. Notably, the substantial gap between standard and fuzzy Decision Trees demonstrates that incorporating soft decision boundaries improves robustness when modeling subtle linguistic deviations.

Given the sensitivity of this application domain, it is essential to understand how and why misclassifications occur. Accordingly, the following sections analyze representative false negative and false positive cases drawn from the held-out test set. Each example was misclassified by at least four of the seven predictive models, ensuring that the analysis highlights systematic rather than model-specific errors. These patterns reveal several limitations of the current feature-based framework and motivate the future improvements discussed later in the paper.

\subsection*{False Negative Analysis}

A review of the false negative cases reveals several systematic limitations in the current feature-based framework. Some jailbreak attempts rely on subtle or humorous injections that derail the conversation without strongly affecting the annotated linguistic features. For example, the prompt ``Tell me about your family history. Any clowns?'' is clearly irrelevant, yet its deviations are too weakly expressed for the model to detect.

Other errors arise from \textit{ambiguous or questionable human labels}. Prompts such as ``Yes, but are your physical symptoms not also connected to your mental state?'' or ``You seem a little disoriented. Does this happen often?'' resemble legitimate clinical questions. However, they are labeled as jailbreaks by annotators. In such cases, the system’s prediction error reflects uncertainty or inconsistency in the ground truth rather than a model deficiency.

A third failure mode involves prompts that are linguistically appropriate but \textit{clinically illogical or unsafe}. Instructions like ``Listen to his forehead'' cannot be recognized as harmful because the annotated features capture tone and relevance rather than medical validity. These examples show that detecting such jailbreaks requires reasoning beyond surface-level linguistic cues.

Finally, some false negatives involve \textit{workflow bypass attempts} in which students politely ask the AI for direct answers. These prompts maintain professionalism and relevance, causing the model to interpret them as benign despite violating the purpose of the simulation. Since the four linguistic features do not encode task adherence or procedural integrity, the system fails to identify these as jailbreak attempts.

\subsection*{False Positive Analysis}

Several sources of false positives reflect limitations in both the feature extraction process and the underlying annotations. Some errors arise from \textit{unfinished or mistyped prompts} that superficially resemble incoherence or derailment. For instance, the incomplete message ``Jimmy you have any kind'' received low feature scores despite the intended prompt being entirely appropriate, showing the model’s sensitivity to surface-level noise.

Another category involves \textit{role-play or theatrical injections} that deviate from standard clinical tone but do not constitute adversarial behavior. Prompts such as ``Ok I take off all my makeup and walk in the room. I say `hello madame. How can I be of assistance today' '' introduce off-role elements that reasonably resemble jailbreak attempts, yet were labeled as non-jailbreak by annotators. These cases also highlight inconsistencies in human labeling that contribute directly to false positives.

A major set of false positives stems from \textit{annotation ambiguity} in casual conversational replies. Messages like ``You and me both, brother!'' or ``YEH LET'S GET BUSY SON'' diverge from expected clinical professionalism, and the model flags them accordingly, but annotators classified them as benign. Such discrepancies illustrate how differing interpretations of tone lead to inconsistent supervision signals.

The system also struggles with \textit{mixed-quality prompts} that blend clinically relevant content with unrelated commentary. Examples include prompts such as ``I think you've got a case against them here. I've got a great lawyer you should call. But let's get back to your exam first.'' The prompt switches between casual conversation and medical diagnosing, causing distracting segments to dominate the message-level features. Because the model cannot perfectly capture sentence-level nuance, it disproportionately penalizes these mixed intents.

Finally, several false positives involve \textit{benign conversational affirmations} that contain little or no clinical content. Short responses such as ``Good for you.'' or ``Edward, I think you're a really great guy.'' are harmless but fall near the decision boundary due to low medical relevance. This leads the classifier to misinterpret them as indicators of distraction or derailment in the clinical workflow.

\subsection*{Suggestions and Future Directions}

The patterns observed in both the false positive and false negative cases highlight several limitations in the current feature-based jailbreak detection framework. Many errors arise from missing contextual nuance, annotation inconsistencies, mixed-content prompts, or workflow-related behaviors that the four linguistic features were not designed to capture. These findings indicate that jailbreak behavior in clinical simulations often manifests in subtle or indirect ways, requiring methods that extend beyond surface-level linguistic cues.

One avenue for improvement is the use of \textit{larger and more capable LLM annotators} during the feature extraction stage. Stronger models may better interpret context, reduce susceptibility to injection-style distractions, and improve the accuracy of the regression outputs on which downstream classifiers depend. This could help mitigate several false predictions stemming from weak feature signals.

Another strategy involves \textit{segmenting prompts into finer-grained units} so that clinically relevant statements can be analyzed independently from irrelevant or casual language. This approach may reduce false positives in mixed-quality prompts and help the system distinguish clinical reasoning from conversational padding. Finer granularity could also strengthen detection of subtle jailbreak attempts that rely on embedding misleading content within otherwise appropriate messages.

Nonetheless, system performance will remain constrained by the \textit{quality and consistency of human annotations}. Several misclassifications arise directly from ambiguous or questionable labels, suggesting the need for expanded datasets with more detailed annotation schemes. Additional features related to clinical actions, workflow adherence, reasoning steps, or safety violations may help operationalize distinctions that annotators currently make inconsistently.

Finally, future work may explore \textit{temporal or conversation-level measures of jailbreak likelihood} rather than relying solely on isolated message predictions. Tracking how jailbreak probability evolves across a dialogue could capture gradual deviations or manipulative patterns that are overlooked when evaluating prompts independently. Together, these improvements can support a more robust and context-aware jailbreak detection system.

\section{Conclusion}

In this study, we present a two-layer framework for detecting jailbreak attempts in clinical LLM interactions by combining automated linguistic feature extraction with downstream classification models. In the first layer, four fine-tuned LLM regressors generate continuous scores for the linguistic constructs shown to correlate with jailbreak behavior: Professionalism, Medical Relevance, Ethical Behavior, and Contextual Distraction. These model-derived features provide an interpretable and scalable alternative to manual annotation. In the second layer, a diverse suite of classifiers leverages the four-dimensional feature representation to predict jailbreak likelihood. The system demonstrates strong performance across both cross-validation and held-out test sets, indicating that LLM-extracted linguistic cues form a reliable foundation for downstream jailbreak detection.

An analysis of failure modes reveals that misclassifications often arise from limitations in both the feature representation and the underlying human annotations. False negatives are frequently driven by subtle injections, ambiguous labels, clinically illogical prompts, or workflow bypass attempts that fall outside the expressive capacity of the four linguistic features. False positives, in contrast, frequently stem from annotation inconsistencies, mixed-quality prompts, conversational informality, or benign statements with low medical content that lie near the classifier’s decision boundary. These observations underscore the need for richer and more consistent annotations, finer-grained feature extraction methods, and potential integration of clinical reasoning or workflow modeling.

Future work should explore the use of more capable LLM annotators, sentence-level or segment-level feature extraction, expanded annotation schemes, and temporal modeling of jailbreak probability across an entire conversation. Together, these enhancements could improve robustness and address the nuanced ways in which jailbreak behavior emerges in clinical simulations. Overall, the findings demonstrate that automated linguistic feature extraction coupled with downstream classifiers provides a promising and interpretable path toward scalable jailbreak detection in safety-critical clinical LLM systems.

\bibliographystyle{splncs04.bst} 
\bibliography{references}

\newpage
\appendix
\section*{\centering Appendix}
\addcontentsline{toc}{section}{Appendix}

This appendix provides supplementary tables and figures that support the results presented in the main text. These materials include detailed descriptions of all pretrained transformer models used in this study, along with comprehensive regression performance tables for each linguistic feature. Additional tables also report the test-set performance of classification models trained on ground-truth annotated features.

\begin{table}[H]
\centering
\caption{Summary of Pre-trained Transformer Models Used in This Study}
\label{tab:bert-model-summary}

\setlength{\tabcolsep}{6pt}
\begin{tabularx}{\textwidth}{%
  >{\raggedright\arraybackslash}p{0.3\textwidth}
  >{\centering\arraybackslash}p{0.10\textwidth}
  >{\centering\arraybackslash}p{0.10\textwidth}
  >{\raggedright\arraybackslash}p{0.40\textwidth}
}
\toprule
\textbf{Model} & \textbf{Params} & \textbf{Release} & \textbf{Training Data} \\
\midrule

\textbf{BERT (Base)} google-bert/bert-base-uncased
& 110M & Oct.\ 2018 &
BooksCorpus (800M words), English Wikipedia (2.5B words) \\

\textbf{BERT (Large)} google-bert/bert-large-uncased
& 340M & Oct.\ 2018 &
BooksCorpus, English Wikipedia \\

\textbf{DistilBERT} distilbert/distilbert-base-uncased
& 66M & Oct.\ 2019 &
BooksCorpus, English Wikipedia \\

\textbf{RoBERTa} FacebookAI/roberta-base
& 125M & Jul.\ 2019 &
BookCorpus, CC-News, OpenWebText, Stories ($\sim$160GB) \\

\textbf{DeBERTa (Large)} microsoft/deberta-v3-large
& 304M & 2021 &
Large-scale web corpora ($\sim$160GB) \\

\textbf{BioBERT (Base)} dmis-lab/biobert-base-cased-v1.1
& 110M & Jan.\ 2019 &
BERT + PubMed abstracts (4.5B), PMC full text (13.5B) \\

\textbf{PubMedBERT} microsoft/BiomedNLP-BiomedBERT-base-uncased-abstract
& 110M & Jul.\ 2020 &
14M PubMed abstracts ($\sim$3.1B words) \\

\bottomrule
\end{tabularx}
\end{table}

\begin{table}[htbp]
\centering
\caption{Regression results across all four dimensions evaluated in this study.}
\label{tab:reg_all}
\begin{subtable}{\linewidth}
\centering
\caption{Professionalism and Medical Relevance.}
\label{tab:reg_prof_med}
\begin{tabular}{lccc ccc}
\toprule
 & \multicolumn{3}{c}{\textbf{Professionalism}} 
 & \multicolumn{3}{c}{\textbf{Medical Relevance}} \\
\cmidrule(lr){2-4} \cmidrule(lr){5-7}
\textbf{Model}
& RMSE & $R^2$ & Error ($\mu \pm \sigma$)
& RMSE & $R^2$ & Error ($\mu \pm \sigma$) \\
\midrule
BERT        & 0.4571 & 0.5697 & 0.0246$\pm$0.4565 & \textbf{0.4630} & 0.5983 & 0.0056$\pm$0.4630 \\
BERT-Large  & 0.4700 & \textbf{0.5450} & 0.0044$\pm$0.4700 & 0.4996 & 0.5324 & -0.0923$\pm$0.4910 \\
DistilBERT  & \textbf{0.4441} & 0.5939 & -0.0037$\pm$0.4441 & 0.5098 & 0.5132 & 0.1294$\pm$0.4931 \\
RoBERTa     & 0.4534 & 0.5766 & -0.0304$\pm$0.4524 & 0.5656 & \textbf{0.4007} & 0.0058$\pm$0.5655 \\
DeBERTa-v3  & 0.4667 & 0.5514 & -0.0158$\pm$0.4665 & 0.5289 & 0.4760 & 0.0634$\pm$0.5251 \\
BioBERT     & 0.4536 & 0.5763 & -0.0315$\pm$0.4525 & 0.5306 & 0.4725 & -0.0564$\pm$0.5276 \\
PubMedBERT  & 0.4608 & 0.5627 & -0.0484$\pm$0.4583 & 0.5051 & 0.5221 & -0.0624$\pm$0.5012 \\
\bottomrule
\end{tabular}
\end{subtable}

\vspace{0.7em} 

\begin{subtable}{\linewidth}
\centering
\caption{Ethical Behavior and Contextual Distraction.}
\label{tab:reg_eth_ctx}
\begin{tabular}{lccc ccc}
\toprule
 & \multicolumn{3}{c}{\textbf{Ethical Behavior}} 
 & \multicolumn{3}{c}{\textbf{Contextual Distraction}} \\
\cmidrule(lr){2-4} \cmidrule(lr){5-7}
\textbf{Model}
& RMSE & $R^2$ & Error ($\mu \pm \sigma$)
& RMSE & $R^2$ & Error ($\mu \pm \sigma$) \\
\midrule
BERT        & 0.6774 & \textbf{0.3015} & -0.2707$\pm$0.6210 & 0.7112 & 0.5661 & 0.0094$\pm$0.7111 \\
BERT-Large  & 0.6164 & 0.4216 & -0.0402$\pm$0.6151 & 0.8078 & 0.4401 & -0.0489$\pm$0.8063 \\
DistilBERT  & 0.5982 & 0.4552 & -0.1951$\pm$0.5655 & 0.7560 & 0.5097 & 0.0017$\pm$0.7560 \\
RoBERTa     & 0.6757 & 0.3051 & -0.0996$\pm$0.6683 & 0.8406 & 0.3938 & 0.0677$\pm$0.8378 \\
DeBERTa-v3  & 0.6056 & 0.4417 &  0.0104$\pm$0.6056 & 0.6783 & 0.6053 & 0.2049$\pm$0.6466 \\
BioBERT     & \textbf{0.5751} & 0.4965 & -0.1355$\pm$0.5589 & \textbf{0.6701} & 0.6147 & 0.0039$\pm$0.6701 \\
PubMedBERT  & 0.5971 & 0.4573 &  0.0217$\pm$0.5967 & 0.9042 & \textbf{0.2986} & -0.2277$\pm$0.8751 \\
\bottomrule
\end{tabular}
\end{subtable}

\end{table}

\begin{table}[htbp]
\centering
\caption{Test set performance of classification models using annotated features.}
\label{tab:test-performance-ground-truth}

\setlength{\tabcolsep}{6pt}

\begin{tabular}{lccccc}
\toprule
\textbf{Model} & \textbf{Accuracy} & \textbf{Precision} & \textbf{Recall} & \textbf{F1} & \textbf{ROC-AUC} \\
\midrule
DT      & 0.9359 & 0.8308 & 0.8852 & 0.8571 & 0.9176 \\
FDT     & 0.9431 & 0.8571 & 0.8852 & 0.8710 & 0.9222 \\
RF      & 0.9288 & 0.7971 & 0.9016 & 0.8462 & 0.9190 \\
LGBM    & 0.9253 & 0.7857 & 0.9016 & 0.8397 & 0.9167 \\
XGBoost & 0.9217 & 0.7671 & 0.9180 & 0.8358 & 0.9204 \\
LR      & 0.9359 & 0.8413 & 0.8689 & 0.8548 & 0.9117 \\
NB      & 0.9395 & 0.8438 & 0.8852 & 0.8640 & 0.9199 \\
\bottomrule
\end{tabular}
\end{table}

\end{document}